\documentclass{article}
\usepackage{ijcai23}

\usepackage{times}
\usepackage{soul}
\usepackage{url}
\usepackage[hidelinks]{hyperref}
\usepackage[utf8]{inputenc}
\usepackage[small]{caption}
\usepackage{graphicx}
\usepackage{amsmath}
\usepackage{amsthm}
\usepackage{booktabs}
\usepackage{algorithm}
\usepackage{algorithmic}
\usepackage[switch]{lineno}
\usepackage{tabularx}
\usepackage[figuresright]{rotating}
\usepackage{graphicx}
\usepackage{tikz}
\usepackage{forest}
\usetikzlibrary{trees,positioning,shapes,shadows,arrows.meta}
\usepackage{xspace}

\usepackage{xcolor}
\definecolor{Cerulean}{RGB}{0, 123, 167} 
\definecolor{OliveGreen}{RGB}{107, 142, 35} 

\usepackage{booktabs}
\usepackage{bbding}

\usepackage{subfigure}

\usepackage{amsmath}
\usepackage{amssymb}
\usepackage{mathtools}
\usepackage{amsthm}

\usepackage{framed,multirow}
\usepackage{lscape}
\usepackage{subcaption}
\usepackage{longtable}
\usepackage{academicons}
\usepackage{threeparttable}
\usepackage{placeins}
\usepackage{float}
\usepackage{xcolor}

\usepackage[utf8]{inputenc}

\title{LLM4Drive: A Survey of Large Language Models for Autonomous Driving}

\author{
Zhenjie Yang$^{1*}$
\and
Xiaosong Jia$^{1*}$\and
Hongyang Li$^{1,2}$\and
Junchi Yan$^{1\dagger}$
\affiliations
$^1$~School of AI and Department of CSE, Shanghai Jiao Tong University \\
$^2$ OpenDriveLab \\
\emails
\{yangzhenjie, jiaxiaosong, hongyangli, yanjunchi\}@sjtu.edu.cn
\\
\normalsize{
$^*$ Equal contributions, order decided by a coin toss  \quad
$^\dagger$Correspondence author}\\
}

\begin{document}

\maketitle

\begin{abstract}
Autonomous driving technology, a catalyst for revolutionizing transportation and urban mobility, has the tend to transition from rule-based systems to data-driven strategies. 
Traditional module-based systems are constrained by cumulative errors among cascaded modules and inflexible pre-set rules. In contrast, end-to-end autonomous driving systems have the potential to avoid error accumulation due to their fully data-driven training process, although they often lack transparency due to their
"black box" nature, complicating the validation and traceability of decisions. Recently, large language models (LLMs) have demonstrated abilities including understanding context, logical reasoning, and generating answers. A natural thought is to utilize these abilities to empower autonomous driving. By combining LLM with foundation vision models, it could open the door to open-world understanding, reasoning, and few-shot learning, which current autonomous driving systems are lacking. In this paper, we 
systematically review the research line about \textit{(Vision) Large Language Models for Autonomous Driving ((V)LLM4Drive)}. This study evaluates the current state of technological advancements, distinctly outlining the principal challenges and prospective directions for the field. For the convenience of researchers in academia and industry, we provide real-time updates on the latest advances in the field as well as relevant open-source resources via the designated link:
\textbf{\textcolor{black}{\url{https://github.com/Thinklab-SJTU/Awesome-LLM4AD}}}.
\end{abstract}

\section{Introduction}

\begin{figure*}
\centering
\includegraphics[width=1\textwidth]{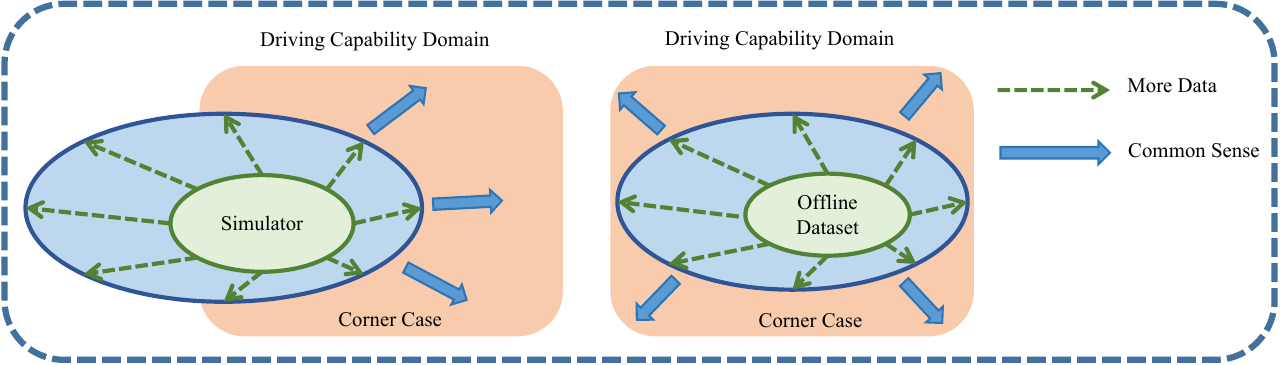}
\caption{The limitation of current autonomous driving paradigm ({\color{OliveGreen}green arrow}) and where LLMs can potentially enhance autonomous driving ability ({\color{Cerulean}blue arrow}).} 
\label{fig:whyllmenhance}
\end{figure*}

Autonomous driving is rapidly reshaping our understanding of transportation, heralding a new era of technological revolution. This transformation means not only the future of transportation but also a fundamental shift across various industries. In conventional autonomous driving systems, algorithms typically adopt the modular design~\cite{liang2020pnpnet,luo2018fast,sadat2020perceive}, with separate components responsible for critical tasks such as perception~\cite{li2022bevformer,liu2022bevfusion}, prediction~\cite{shi2022motion,jia2022temporal,jia2023hdgt,jia2024amp}, and planning~\cite{treiber2000congested,Dauner2023CORL,li2024think,jia2024bench}. Specifically, the perception component handles object detection~\cite{li2022bevformer,liu2022bevfusion}, tracking~\cite{zeng2021motr}, and sophisticated semantic segmentation tasks~\cite{cheng2021mask2former}. The prediction component analyzes the external environment~\cite{jia2021ide} and estimates the future states of the surrounding agents~\cite{jia2022multi}. The planning component, often reliant on rule-based decision algorithms~\cite{treiber2000congested}, determines the optimal and safest route to a predetermined destination. While the module-based approach provides reliability and enhanced security in a variety of scenarios, it also presents challenges. The decoupled design between system components may lead to key information loss during transitions and potentially redundant computation as well. Additionally, errors may accumulate within the system due to inconsistencies in optimization objectives among the modules, affecting the vehicle's overall decision-making performance~\cite{chen2023end}.

Rule-based decision systems, with their inherent limitations and scalability issues, are gradually giving way to data-driven methods. End-to-end autonomous driving solutions are increasingly becoming a consensus in the field~\cite{wu2022trajectoryguided,Chitta2023PAMI,chen2022lav,jia2023think,jia2023driveadapter,hu2023planning}. By eliminating integration errors between multiple modules and reducing redundant computations, the end-to-end system enhances the expression of visual~\cite{wu2022policy} and sensory information while ensuring greater efficiency. However, this approach also introduces the ``black box" problem, meaning a lack of transparency in the decision-making process, complicating interpretation and validation.

Simultaneously, the explainability of autonomous driving has become an important research focus \cite{jin2023adapt}. Although smaller language models (like early versions of BERT \cite{devlin2018bert} and GPT \cite{brown2020language}) employed in massive data collection from driving scenarios help address this issue, they often lack sufficient generalization capabilities to perform optimally. Recently, large language models \cite{openai2023gpt4,touvron2023llama} have demonstrated remarkable abilities in understanding context, generating answers, and handling complex tasks. They are also now integrated with multimodal models \cite{brohan2023rt,liu2023visual,driess2023palme,xu2023drivegpt4,chen2023driving}. This integration achieves a unified feature space mapping for images, text, videos, point clouds, etc. Such consolidation significantly enhances the system's generalization capabilities and equips them with the capacity to quickly adapt to new scenarios in a zero-shot or few-shot manner.

In this context, developing an interpretable and efficient end-to-end autonomous driving system has become a research hotspot~\cite{chen2023end}. Large language models, with their extensive knowledge base and exceptional generalization, could facilitate easier learning of complex driving behaviors. By leveraging the visual-language model (VLM)'s robust and comprehensive capabilities of open-world understanding and in-context learning~\cite{bommasani2021opportunities,brohan2023rt2,liu2023visual,driess2023palme}, it becomes possible to address the long-tail problem for perception networks, assist in decision-making, and provide intuitive explanations for these decisions. 

This paper aims to provide a comprehensive overview of this rapidly emerging research field, analyze its basic principles, methods, and implementation processes, and introduce in detail regarding the application of LLMs for autonomous driving. Finally, we discuss related challenges and future research directions.

\section{Motivation of LLM4AD}

In today's technological landscape, large language models such as GPT-4 and GPT-4V~\cite{openai2023gpt4,yang2023dawn} are drawing attention with their superior contextual understanding and in-context learning capabilities. Their enriched common sense knowledge has facilitated significant advancements in many downstream tasks. We ask the question: \textit{how do these large models assist in the domain of autonomous driving, especially in playing a critical role in the decision-making process?}

In Fig.~\ref{fig:whyllmenhance}, we give an intuitive demonstration of the limitation of current autonomous driving paradigm and where LLMs can potentially enhance autonomous driving ability. We summarize two primary aspects of driving skills. The orange circle represents the ideal level of driving competence, akin to that possessed by an experienced human driver. There are two main methods to acquire such proficiency: one, through learning-based techniques within simulated environments; and two, by learning from offline data through similar methodologies. It's important to note that due to discrepancies between simulations and the real world, these two domains are not fully the same, i.e. sim2real gap~\cite{hofer2021sim2real}. Concurrently, offline data serves as a subset of real-world data since it's collected directly from actual surroundings. However, it is difficult to fully cover the distribution as well due to the notorious long-tailed nature~\cite{jain2021autonomy} of autonomous driving tasks.

The final goal of autonomous driving is to elevate driving abilities from a basic green stage to a more advanced blue level through extensive data collection and deep learning. However, the high cost associated with data gathering and annotation, along with the inherent differences between simulated and real-world environments, mean there's still a gap before reaching the expert level of driving skills. In this scenario, if we can effectively utilize the innate common sense embedded within large language models, we might gradually narrow this gap. Intuitively, by adopting this approach, we could progressively enhance the capabilities of autonomous driving systems, bringing them closer to, or potentially reaching, the ideal expert level of driving proficiency. Through such technological integration and innovation, we anticipate significant improvements in the overall performance and safety of autonomous driving.

\begin{figure*}
\centering
\includegraphics[width=0.95\textwidth]{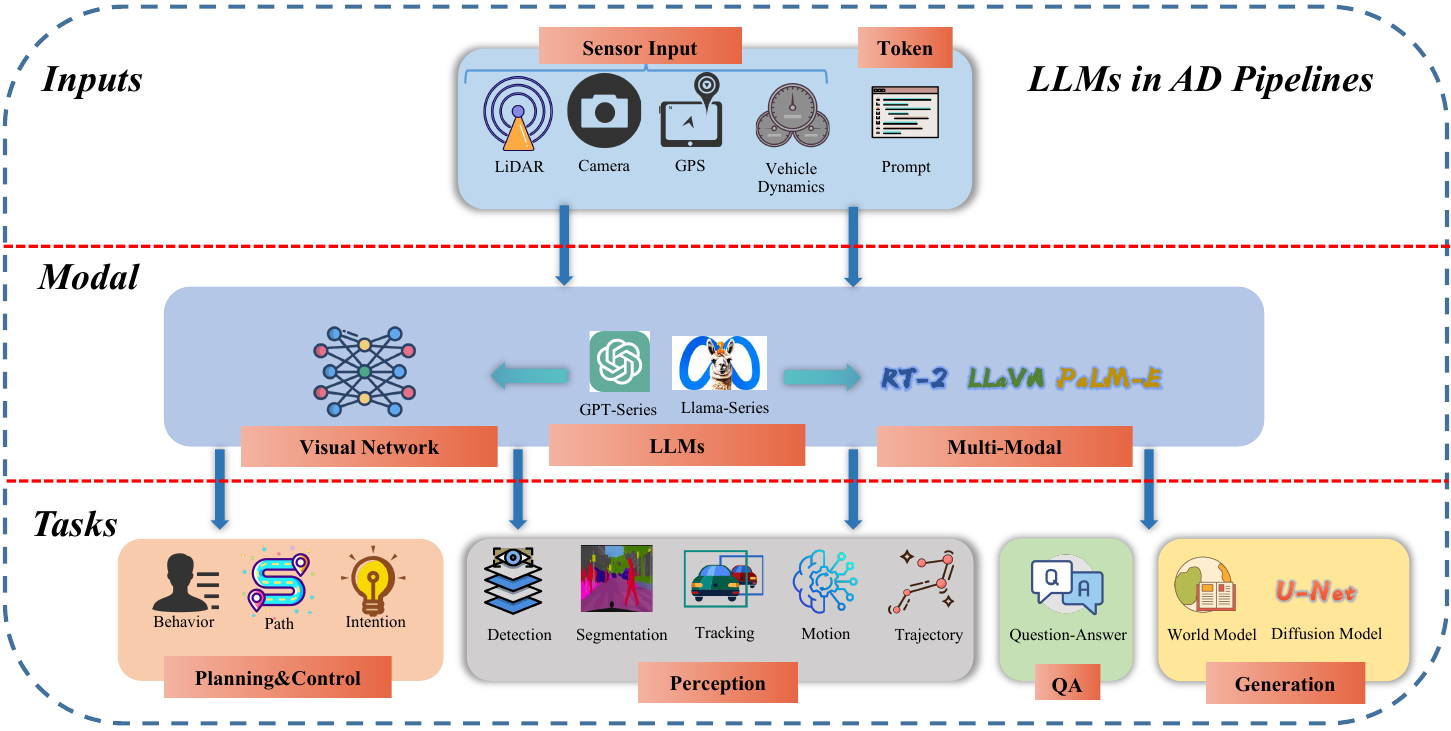}
\caption{LLMs in Autonomous Driving Pipelines.}
\label{fig:llm4adpipeline}
\end{figure*}

The application of large language models in the field of autonomous driving indeed covers a wide range of task types, combining depth and breadth with revolutionary potential. LLMs in autonomous driving pipelines is shown in the Fig.~\ref{fig:llm4adpipeline}.  

\begin{figure*}
  \centering
  \begin{forest}
  for tree={
    grow=east, 
    parent anchor=east,
    child anchor=west,
    edge={cyan, line width=1pt},
    edge path={
      \noexpand\path [\forestoption{edge}] (!u.parent anchor) -- ++(5pt,0) |- (.child anchor)\forestoption{edge label};
    }
    rounded corners, 
    draw, 
    align=center, 
    font=\scriptsize,
    minimum size=3mm,
    for children={
    }
    }
    [LLM4AD, fill=red!50, 
        [Evaluation \& Benchmark, fill=cyan!40 
            [
            On the Road with GPT-4V \cite{wen2023road} \\ 
            GPT-4V Takes the Wheel \cite{huang2023gpt4v} \\ 
            LaMPilot \cite{ma2023lampilot} \\
            Evaluation of LLMs \cite{tanahashi2023evaluation} \\
            Testing LLMs \cite{tang2024testing} \\
            DriveSim \cite{sreeram2024probing} \\
            ELM \cite{zhou2024embodied} \\
            LimSim++ \cite{fu2024limsim} \\
            OmniDrive \cite{wang2024omnidrive} \\
            AIDE \cite{liang2024aide}
            ]
        ]
       [Question Answering(QA), fill=orange!70
            [Traditional QA, fill=orange!70
                [Domain Knowledge Distillation \cite{tang2023domain} \\
                Human-Centric Autonomous Systems \cite{yang2023humancentric} \\
                Engineering Safety \cite{nouri2024engineeringsafetyrequirementsautonomous} \\
                Hybrid Reasoning \cite{azarafza2024hybrid}
                ]
            ]
            [Visual QA, fill=orange!70
                [DriveMLM \cite{wang2023drivemlm} \\
                DriveLM \cite{sima2023drivelm} \\
                Reason2Drive \cite{nie2023reason2drive} \\
                LingoQA \cite{marcu2023lingoqa} \\
                Dolphins \cite{ma2023dolphins} \\
                DriveGPT4 \cite{xu2023drivegpt4} \\
                A Superalignment Framework \cite{kong2024superalignment} \\
                EM-VLM4AD \cite{gopalkrishnan2024multi} \\
                TransGPT \cite{wang2024transgptmultimodalgenerativepretrained}
                ]
            ]
        ]
        [Generation(Diffusion), fill=yellow!70, l sep+=1em
                [ADriver-I \cite{jia2023adriveri} \\ 
                DrivingDiffusion \cite{li2023drivingdiffusion} \\
                DriveDreamer \cite{wang2023drivedreamer} \\
                CTG++ \cite{zhong2023languageguided} \\
                GAIA-1 \cite{hu2023gaia} \\
                MagicDrive \cite{gao2023magicdrive} \\
                Driving into the Future \cite{wang2023driving} \\
                ChatScene \cite{zhang2024chatscene} \\
                REvolve \cite{hazra2024revolve} \\
                GenAD \cite{yang2024generalized} \\
                DriveDreamer-2 \cite{zhao2024drivedreamer} \\
                ChatSim \cite{wei2024editable} \\
                LLM-Assisted Light \cite{wang2024llmassistedlightleveraginglarge} \\
                LangProp \cite{ishida2024langprop}
                ]
        ]
        [Planning \& Control, fill=blue!30, name=rightNode
            [Fine-tuning Pre-trained Model, fill=blue!30
                [DriveMLM \cite{wang2023drivemlm} \\
                LMDriver \cite{shao2023lmdrive} \\
                Agent-Driver \cite{mao2023language} \\
                GPT-Driver \cite{mao2023gpt} \\
                DriveLM \cite{sima2023drivelm} \\
                DriveGPT4 \cite{xu2023drivegpt4} \\
                Driving with LLMs \cite{chen2023driving} \\
                MTD-GPT \cite{liu2023mtdgpt} \\
                KoMA \cite{jiang2024koma} \\
                AsyncDriver \cite{chen2024asynchronous} \\
                PlanAgent \cite{zheng2024planagent} \\
                AgentsCoDriver \cite{hu2024agentscodriver} \\
                DriveVLM \cite{tian2024drivevlm} \\
                RAG-Driver \cite{yuan2024rag} \\
                VLP \cite{pan2024vlp} \\
                DME-Driver \cite{han2024dme} \\
                ]
            ]
            [Prompt Engineering, fill=blue!30
                [A Safety Perspective \cite{wang2023empowering} \\ 
                Talk2Drive \cite{cui2023large} \\
                ChatGPT as Your Vehicle Co-Pilot \cite{10286969} \\
                Receive Reason and React \cite{cui2023receive} \\
                LanguageMPC \cite{sha2023languagempc} \\
                Talk2BEV \cite{dewangan2023talk2bev} \\
                SurrealDriver \cite{jin2023surrealdriver} \\
                Drive as You Speak \cite{cui2023drive} \\
                TrafficGPT \cite{zhang2023trafficgpt} \\
                Drive Like a Human \cite{fu2023drive} \\
                DiLu \cite{wen2023dilu} \\
                LLM-assited light \cite{wang2024llm} \\
                AccidentGPT \cite{wang2023accidentgpt} \\
                LLM-Assist \cite{sharan2023llm} \\
                LLaDA \cite{Li_2024_CVPR}
                ]
            ]
        ]
        [Perception, fill=green!50
            [Prediction, fill=green!50
                 [Can you text what is happening \cite{keysan2023text} \\
                 MTD-GPT \cite{liu2023mtdgpt} \\
                 LC-LLM \cite{peng2024lc} \\
                 LeGo-Drive \cite{paul2024lego} \\
                 Context-aware Motion Prediction \cite{zheng2024large}
                 ]
            ]
            [Detection, fill=green!50
                 [HiLM-D \cite{ding2023hilmd}
                 ]
            ]
            [Tracking, fill=green!50
                 [LanguagePrompt \cite{wu2023language}]
            ]
        ]
    ]
    \end{forest}
  \caption{Large Language Models for Autonomous Driving Research Tree}
  \label{fig:Research_Tree}
\end{figure*}
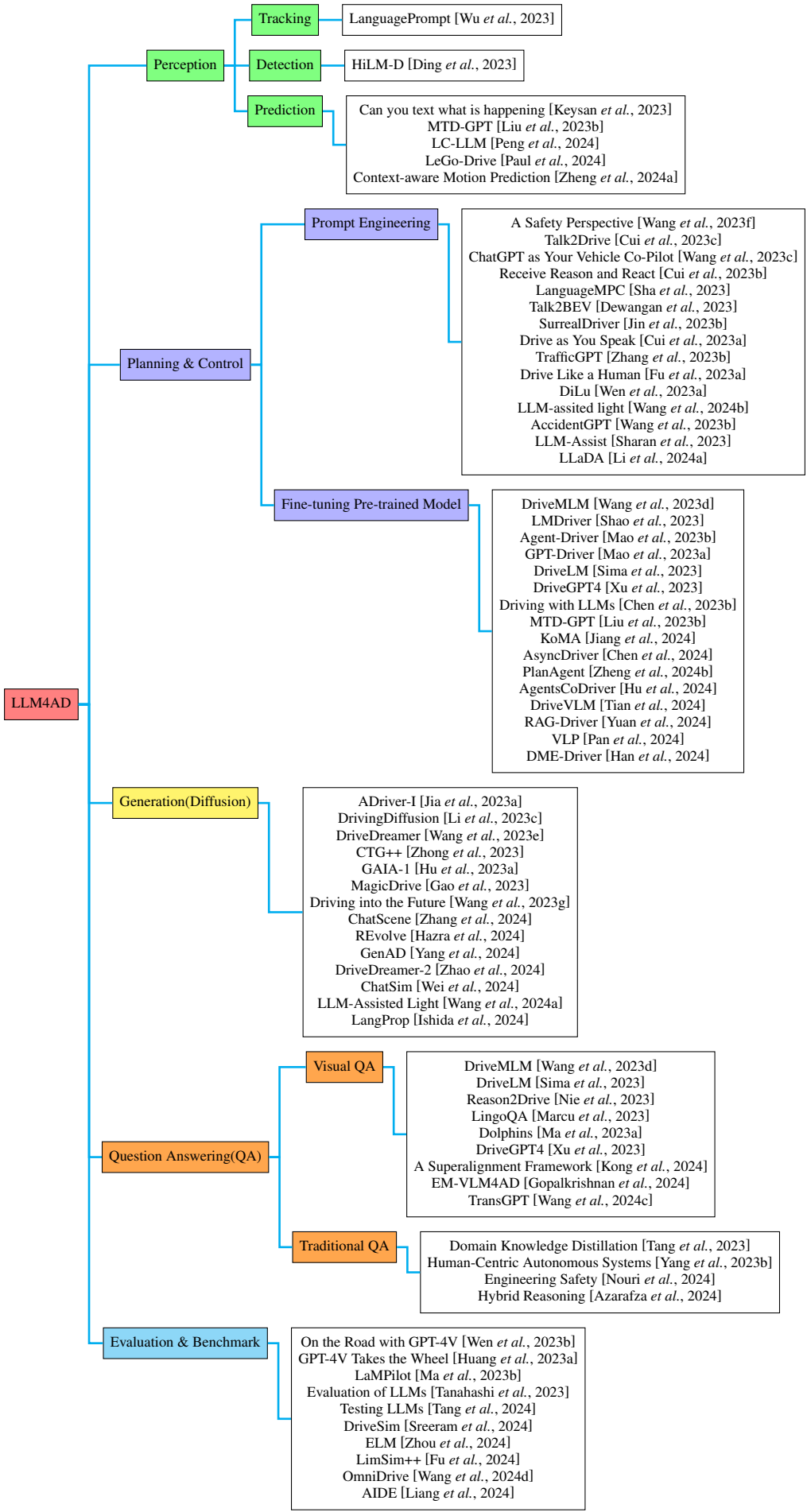

\section{Application of LLM4AD}
In the following sections, we divide existing works based on the perspective of applying LLMs: planning, perception, question answering, and generation. The corresponding taxonomy tree is shown in Fig.~\ref{fig:Research_Tree}.

\subsection{Planning \& Control}
Large language models (LLMs) have achieved great success with their open-world cognitive and reasoning capabilities~\cite{radford2018improving,radford2019language,brown2020language,ouyang2022training,openai2023gpt4}. These capabilities could provide a transparent explanation of the autonomous driving decision-making process, significantly enhancing system reliability and user trust in the technology~\cite{Deruyttere_2019,kim2019grounding,atakishiyev2023explainable,jin2023adapt,malla2023drama}. Within this domain, based on whether tuning the LLM, related research can be categorized into two main types: fine-tuning pre-trained models and prompt engineering.

\subsubsection{Fine-tuning pre-trained models}
In the application of fine-tuning pre-trained models, MTD-GPT ~\cite{liu2023mtdgpt} translates multi-task decision-making problems into sequence modeling problems. Through training on a mixed multi-task dataset, it addresses various decision-making tasks at unsignaled intersections. Although this approach outperforms the performance of single-task decision-making RL models, the used scenes are limited to unsignaled intersections, which might be enough to demonstrate the complexity of the real world application. Driving with LLMs ~\cite{chen2023driving} designs an architecture that fuses vectorized inputs into LLMs with a two-stage pretraining and fine-tuning method. Due to the limitation of vectorized representations, their method are only tested in the simulation. DriveGPT4~\cite{xu2023drivegpt4} presents a multimodal LLM based on Valley~\cite{luo2023valley} and develops a visual instruction tuning dataset for interpretable autonomous driving. Besides predicting a vehicle's basic control signals, it also responds in real-time, explaining why the action is taken. It outperforms baseline models in a variety of QA tasks while the experiments about planning is simple. GPT-Driver~\cite{mao2023gpt} transforms the motion planning task into a language modeling problem. It exceeds the UniAD\cite{hu2023planning} in the L2 metric. Nevertheless, since it uses past speed and acceleration information, there is concern about unfair comparison with UniAD. Additionally, L2 only reflects the fitting degree of the driving route and might not reflect the driving performance~\cite{Dauner2023CORL}. Agent-Driver~\cite{mao2023language} leverages LLMs common sense and robust reasoning capabilities to improve the capabilities of planning by designing a tool library, a cognitive memory, and a reasoning engine. This paradigm achieves better results on the nuScenes dataset. Meanwhile, shortening the inference time is also an urgent problem. DriveLM~\cite{sima2023drivelm} uses a trajectory tokenizer to process ego-trajectory signals to texts, making them belong to the same domain space. Such a tokenizer can be applied to any general vision language models. Moreover, they utilize a graph-structure inference with multiple QA pairs in logical order, thus improving the final planning performance. \cite{10286969} adapts LLMs as a vehicle "Co-Pilot" of driving, which can accomplish specific driving tasks with human intention satisfied based on the information provided. It lacks verification in complex interaction scenarios. LMDrive~\cite{shao2023lmdrive} designs a multi-modal framework to predict the control signal and whether the given instruction is completed. It adopts Resnet~\cite{He_2016_CVPR} as the vision encoder which has not been through an image-text alignment pretraining. In addition, it introduces a benchmark LangAuto which includes approximately 64K instruction-following data clips in CARLA. The
LangAuto benchmark tests the system’s ability to handle complex instructions and challenging driving scenario. DriveMLM~\cite{wang2023drivemlm} adopts a multi-modal LLM(Multi-view image, Point cloud, and prompt) to generate high-level decision commands and uses Apollo as a planner to get the control signal. Moreover, the training data generated by experts and uses GPT-3.5 to increase data diversity. It achieves 76.1 driving score on the CARLA Town05 Long, which reaches the level of classic end-to-end autonomous driving. KoMA~\cite{jiang2024koma} is a knowledge-driven multi-agent framework in which each agent is powered by large language models. These agents analyze and infer the intentions of surrounding vehicles to enhance decision-making. AsyncDriver~\cite{chen2024asynchronous} is an asynchronous LLM-enhanced framework where the inference frequency of LLM is controllable and can be decoupled from the real-time planner. It has good closed-loop evaluation performance in challenging scenarios of nuPlan. PlanAgent~\cite{zheng2024planagent} extracts bird's-eye view (BEV) representation and generates a text description input based on the lane map through an environment transformation module. It uses a reasoning engine module to perform a hierarchical thinking chain to guide driving scene understanding, motion command generation, and planning code writing. AGENTSCODRIVER~\cite{hu2024agentscodriver} is an LLM-powered framework for multi-vehicle collaborative driving with lifelong learning, enabling communication and collaboration among driving agents in complex traffic scenarios, featuring a reasoning engine, cognitive memory, reinforcement reflection, and a communication module. DriveVLM~\cite{tian2024drivevlm} leverages Vision-Language Models to enhance scene understanding and planning capabilities for autonomous driving, while DriveVLM-Dual synergizes these advancements with traditional 3D perception and planning approaches to effectively address spatial reasoning and computational challenges, demonstrating superior performance in complex and dynamic driving scenarios. RAG-Driver~\cite{yuan2024rag}, a Multi-Modal Large Language Model with Retrieval-augmented In-context Learning, provides explainable and generalizable end-to-end driving by producing numerical control signals, along with explanations and justifications for driving actions, and demonstrates impressive zero-shot generalization to unseen environments without additional training. LLaDA~\cite{Li_2024_CVPR} designs a training-free mechanism to assist human drivers and adapt autonomous driving policies to new environments. VLP~\cite{pan2024vlp} is a Vision-Language-Planning model intended to enhance autonomous driving systems (ADS) by incorporating two novel components: ALP and SLP. ALP (Agent-wise Learning Paradigm) aligns the generated bird’s-eye-view (BEV) with the true BEV map, improving self-driving BEV reasoning. SLP (Self-Driving-Car-Centric Learning Paradigm) aligns the ego vehicle’s query features with its textual planning features, enhancing self-driving decision-making. DME-Driver~\cite{han2024dme} enhances decision logic explainability and environmental perception accuracy by using a vision language model for decision-making and a planning-oriented perception model for generating precise control signals, effectively translating human-like driving logic into actionable commands, and achieving high-precision planning accuracy through the comprehensive HBD dataset.

\subsubsection{prompt engineering}
In the prompt engineering perspective, some methods tried to tap into the deep reasoning potential of the LLMs through clever prompt design. DiLu~\cite{wen2023dilu} designs a framework of LLMs as agents to solve closed-loop driving tasks. This method introduces a memory module to record experience, to leverage LLMs to facilitate reasoning and reflection processes. DiLu exhibits strong generalization capabilities compared with SOTA RL-based methods. However, the reasoning and reflection processes require multiple rounds of question-answering, and its inference time cannot be ignored. Similarly, Receive Reason and React ~\cite{cui2023receive} and Drive as You Speak~\cite{cui2023drive} integrate the language and reasoning capabilities of LLMs into autonomous vehicles. In addition to memory and reflection processes, these methods introduce additional raw sensor information such as camera, GNSS, lidar, and radar. However, the inference speed is unsolved as well. Furthermore, SurrealDriver~\cite{jin2023surrealdriver} divides the memory module into short-term memory, long-term guidelines, and safety criteria. Meanwhile, it interviews 24 drivers and uses their detailed descriptions of driving behaviors as chain-of-thought prompts to develop a `coach agent' module. However, there is a lack of comparison with traditional algorithms to prove that large language models indeed bring performance improvements. LanguageMPC~\cite{sha2023languagempc} also designs a chain-of-thought framework for LLMs in driving scenarios and it integrates with low-level controllers by guided parameter matrix adaptation. Although its performance exceeds MPC and RL-based methods in the simplified simulator environments, it lacks validation in complex environments. TrafficGPT~\cite{zhang2023trafficgpt} is a fusion of ChatGPT and traffic foundation models which can tackle complex traffic-related problems and provide insightful suggestions. It leverages multimodal data as a data source, offering comprehensive support for various traffic-related tasks. Talk2BEV~\cite{dewangan2023talk2bev} introduces a large vision-language model (LVLM) interface for bird’s-eye view (BEV) maps in autonomous driving contexts. It does not require any training or fine-tuning, only relying on pre-trained image-language models. In addition, it presents a benchmark for evaluating subsequent work in LVLMs for AD applications. Talk2Drive~\cite{cui2023drive} utilizes human verbal commands and makes autonomous driving decisions based on contextual information to meet humanly personalized preferences for safety, efficiency, and comfort. AccidentGPT ~\cite{wang2023accidentgpt} integrates multi-vehicle collaborative perception to improve environmental understanding and collision avoidance, offering advanced safety features like proactive remote safety warnings and blind spot alerts. It also supports traffic police and management agencies by providing real-time intelligent analysis of traffic safety factors.

\textbf{Metric:}

MTD-GPT ~\cite{liu2023mtdgpt} uses single-subtask success rates as the metric in simulation and it exceeds RL expert. DriveGPT4~\cite{xu2023drivegpt4} and RAG-Driver~\cite{yuan2024rag} uses root mean
squared error (RMSE) and threshold accuracies for evaluation. In vehicle action description, justification, and full sentences, it uses BLEU-4~\cite{papineni2002bleu}, METEOR~\cite{banerjee-lavie-2005-meteor}, CIDER\cite{vedantam2015cider} and chatgpt score~\cite{fu2023gptscore}. Driving with LLMs ~\cite{chen2023driving} uses the Mean Absolute Error (MAE) for the predictions of the number of cars and pedestrians, normalized acceleration, and brake pressure. Additionally, it measures the accuracy of traffic light detection as well as the mean absolute distance error in meters for traffic light distance prediction. Besides perception-related metrics, it also uses GPT-3.5 to grade their model’s answers which is a recently emerging technique - grading natural language responses ~\cite{fu2023gptscore,wang2023chatgpt,liu2023geval}. DiLu~\cite{wen2023dilu} uses Success Steps in simulation as a metric to evaluate generalization and transformation abilities. SurrealDriver~\cite{jin2023surrealdriver} evaluates agents based on two main dimensions: safe driving ability and humanness. Safe driving capabilities are assessed through collision rates, while human likeness is assessed through user experiments with 24 adult participants (age 29.3 ± 4.9 years, male = 17 years) who are legal to drive. LanguageMPC~\cite{sha2023languagempc} customs some metrics: failure/collision cases, the efficiency of traffic flow, time cost by ego vehicle, and the safety of the ego vehicle’s driving behavior.
Similarly, Talk2BEV~\cite{dewangan2023talk2bev} measures their methods from the perspective of spatial
reasoning, instance attribute, instance counting, and visual reasoning.
GPT-Driver~\cite{mao2023gpt}, LLaDA~\cite{Li_2024_CVPR}, DriveLM~\cite{sima2023drivelm}, Agent-Driver~\cite{mao2023language}, VLP~\cite{pan2024vlp}, DME-Driver~\cite{han2024dme} and DriveVLM~\cite{tian2024drivevlm} contain two metrics: L2 error (in meters) and collision rate (in percentage). The average L2 error is calculated by measuring the distance of each waypoint in the planned trajectory and the offline recorded human driver trajectory. It reflects the fitting of the planned trajectory to the human driving trajectory. The collision rate is calculated by placing an ego vehicle box at each waypoint of the planned trajectory and then checking for collisions with the ground truth bounding boxes of other objects. It reflects the safety of the planned trajectory. LMDrive~\cite{shao2023lmdrive} and DriveMLM\cite{wang2023drivemlm} adopts CARLA's official metrics including Driving Score(DS), Route Completion(RC), Infraction Score(IS).
At present, LLM4AD regarding the planning task lacks a unified metric and cannot uniformly evaluate the pros and cons between each method and traditional counterparts. KoMA~\cite{jiang2024koma} demonstrates the effectiveness and high success rate in the Highway MARL simulator. AsyncDriver~\cite{chen2024asynchronous} obtains superior closed-loop evaluation performance in nuPlan Closed-Loop Reactive Hard20 scenarios. PlanAgent~\cite{zheng2024planagent} achieves competitive and general results on the nuPlan Val14 and Test14-hard benchmarks, and improves the efficiency of token usage when describing driving scenarios. AGENTSCODRIVER~\cite{hu2024agentscodriver} adopt Success
Rate (SR) and Success Step (SS) in HighwayEnv simulator~\cite{highway-env}. LeGo-Drive~\cite{paul2024lego} is a novel planning-guided end-to-end LLM-based goal point navigation solution that predicts and improves the desired state by dynamically interacting with the environment and generating a collision-free trajectory.  

\subsection{Perception}

Large language models have demonstrated their unique value and strong capabilities in ``perception" tasks~\cite{radford2021learning,li2022blip,li2023blip2,li2023toponet,li2022bevsurvey}. Especially in environments where data is relatively scarce, these models can rely on their few-shot learning characteristics to achieve fast and accurate learning and reasoning~\cite{p2023protoclip,lin2023multimodality}. This learning ability is of significance in the perception stage of the autonomous driving system, which greatly improves the system's adaptability and generalization capabilities in changing and complex driving environments. PromptTrack~\cite{wu2023language} fuses cross-modal features in a prompt reasoning branch to predict 3D objects. It uses language prompts as semantic cues and combines LLMs with 3D detection tasks and tracking tasks. Although it achieves better performance compared to other methods, the advantages of LLMs do not directly affect the tracking task. Rather, the tracking task serves as a query to assist LLMs in performing 3D detection tasks. HiLM-D~\cite{ding2023hilmd} incorporates high-resolution information into multimodal large language models for the Risk Object Localization and Intention and Suggestion Prediction (ROLISP) task. It combines LLMs with 2D detection tasks and obtains better performance in detection tasks and QA tasks compared to other multi-modal large models such as Video-LLaMa~\cite{damonlpsg2023videollama}, eP-ALM~\cite{shukor2023epalm}. It is worth noting to point out one potential limitation of the dataset: each video contains only one risk object, which might not capture the complexity of real-world scenarios. \cite{keysan2023text} integrates pre-trained language models as text-based input encoders for the autonomous driving trajectory prediction task. Joint encoders(image and text) over both modalities perform better than using a single encoder in isolation. While the joint model significantly improves the baseline, its performance has not reached the state-of-the-art level yet~\cite{deo2021multimodal,gilles2021gohome}. LC-LLM~\cite{peng2024lc} is designed for lane change prediction, leveraging LLM capabilities to understand complex scenarios, enhancing prediction performance, and providing explainable predictions by generating explanations for lane change intentions and trajectories. AIDE~\cite{liang2024aide} introduces a paradigm for an automatic data engine, incorporating automatic data querying and labeling using VLM, and continual learning with pseudo labels. It introduces a new benchmark to evaluate such automated data engines for AV perception that allows combined insights across multiple paradigms of open vocabulary detection, semi-supervised, and continual learning.          Context-aware Motion Prediction~\cite{zheng2024large} designs and conducts prompt engineering to enable GPT4-V to comprehend complex traffic scenarios. It combines the context information outputted by GPT4-V with MTR~\cite{shi2023motiontransformerglobalintention} to enhance motion prediction.

\textbf{Metric:}

PromptTrack~\cite{wu2023language} uses the Average Multiple Object Tracking Precision
(AMOTA) metric~\cite{journals/ejivp/BernardinS08}, the Average MultiObject Tracking Precision (AMOTP)~\cite{bashar2022multiple} and Identity Switches (IDS)~\cite{huang2023detection} metrics. 
HiLM-D~\cite{ding2023hilmd} uses the BLEU-4~\cite{papineni2002bleu}, METEOR~\cite{banerjee-lavie-2005-meteor}, CIDER\cite{vedantam2015cider} and SPICE~\cite{anderson2016spice}, IoU~\cite{rezatofighi2019generalized} as metrics to compare with the state-of-the-art. ~\cite{keysan2023text} uses the standard evaluation metrics that are provided in the nuScenes-devkit ~\cite{nuscenes2019,fong2021panoptic}: minimum Average Displacement Error (minADEk), Final Displacement Error (minFDEk), and the miss rate over 2 meters. LC-LLM \cite{peng2024lc} uses RMSE (lat) to assess lateral and longitudinal prediction error. LeGo-Drive~\cite{paul2024lego} adopts minFDE (minimum Final Displacement Error) and L2 distance between the goal location and the trajectory endpoint. as evaluation metrics. Context-aware Motion Prediction~\cite{zheng2024large} use the mean Average Precision (mAP) in official WOMD~\cite{ettinger2021largescaleinteractivemotion} evaluation.

\begin{table*}[!ht]
\newcommand{\tabincell}[2]{\begin{tabular}{@{}#1@{}}#2\end{tabular}}
\centering
\begin{tabular}{|l|l|l|l|l|}
\hline
\textbf{Dataset} & \textbf{Task} & \textbf{Size} & \textbf{Annotator} & \textbf{Description} \\
\hline
\tabincell{l}{BDD-X \\ \cite{kim2018textual}} & \tabincell{l}{Planning \\ VQA} &\tabincell{l}{77 hours, 6970 videos, \\ 8.4M frames, 26228 captions} & \tabincell{l}{Human} & \tabincell{l}{Ego-vehicle actions description and\\ explanation.} \\
\hline
\tabincell{l}{HAD \\ \cite{kim2019CVPR}} & \tabincell{l}{Planning \\ Perception} & \tabincell{l}{30 hours, 5744 videos \\ 22366 captions} & \tabincell{l}{Human} & \tabincell{l}{Joint action description for goal-\\oriented advice and attention description\\for stimulus-driven advice.} \\
\hline
\tabincell{l}{Talk2Car \\ \cite{Deruyttere_2019}} & \tabincell{l}{Planning \\ Perception} & \tabincell{l}{15 hours, 850 videos of 20s each \\ 30k frames, 11959 captions} & \tabincell{l}{Human} & \tabincell{l}{Object referral dataset that contains \\ commands written in natural language \\ for self-driving cars.} \\
\hline
\tabincell{l}{DriveLM \\ \cite{sima2023drivelm}} & \tabincell{l}{Perception \\ Prediction \\ Planning \\ VQA} & \tabincell{l}{In Carla, 18k frames and 3.7M\\QA pairs;In nuScenes, 4.8k\\frames and 450k QA pairs} & \tabincell{l}{Human \\ Rule-Based} & \tabincell{l}{P3 with reasoning logic; Connect the \\ QA pairs in a graph-style structure; \\Use ``What if"-style questions.} \\
\hline
\tabincell{l}{DRAMA \\ \cite{malla2023drama}} & \tabincell{l}{VQA} & \tabincell{l}{91 hours, 17785 videos, 77639 \\ question, 102830 answering, \\ 17066 captions} & \tabincell{l}{Human} & \tabincell{l}{Joint risk localization with visual \\reasoning of driving risks in a free-\\form language description.} \\
\hline
\tabincell{l}{Rank2Tell \\ \cite{sachdeva2023rank2tell}} & \tabincell{l}{Perception \\ VQA} & \tabincell{l}{several hours, 118 videos of 20s \\ each} & \tabincell{l}{Human} & \tabincell{l}{Joint important object identification,\\important object localization ranking, \\ and reasoning.} \\
\hline
\tabincell{l}{NuPrompt \\ \cite{wu2023language}} & Perception & \tabincell{l}{15 hours, 35367 prompts for 3D \\ objects} & \tabincell{l}{Huamn \\  GPT3.5} & 
\tabincell{l}{Object-centric language prompt set \\ for perception tasks.} \\
\hline
\tabincell{l}{NuScenes-QA \\ \cite{qian2023nuscenes}} & VQA & \tabincell{l}{15 hours, Train(24149  scences, \\459941 QA pairs), Test(6019 \\ scences, 83337 QA pairs)} & \tabincell{l}{Rule-Based} & 
\tabincell{l}{Leverage 3D annotations(object \\category, position, orientation, \\ relationships information) and designed \\ question templates to construct QA pairs.} \\
\hline
\tabincell{l}{Reason2Drive \\ \cite{nie2023reason2drive}} & \tabincell{l}{Perception \\Prediction\\ VQA}  & \tabincell{l}{600K video-text pairs} & \tabincell{l}{Human \\ GPT-4} & 
\tabincell{l}{Composed of nuScenes, Waymo and \\ ONCE, with driving instructions.} \\
\hline
\tabincell{l}{LingoQA \\ \cite{nie2023reason2drive}} & \tabincell{l}{VQA}  & \tabincell{l}{419.9k QA pairs, 28k scenarios} & \tabincell{l}{Rule-Based \\ GPT-3.5/4 \\ Software \\ Human} & 
\tabincell{l}{Contains reasoning pairs in addition to \\ object presence, description, and \\ localisation. } \\
\hline
\tabincell{l}{NuInstruct \\ \cite{ding2024holistic}} & \tabincell{l}{Perception \\Prediction\\ VQA}  & \tabincell{l}{91k QA pairs, 17 subtasks} & \tabincell{l}{GPT-4 \\ Human} & 
\tabincell{l}{Integrates multi-view information, \\ requiring responses from multiple \\ perspectives, with balanced view \\ distribution for perception tasks.} \\
\hline
\tabincell{l}{OpenDV-2K \\ \cite{yang2024generalized}} & \tabincell{l}{Perception \\Prediction\\ VQA}  & \tabincell{l}{2059 hours of videos paired \\ with texts(1747 hours from \\ YouTube and 312 hours from \\ public datasets). } & \tabincell{l}{BLIP-2} & 
\tabincell{l}{A large-scale multimodal dataset \\ for autonomous driving, to support \\ the training of a generalized video \\ prediction model.} \\

\hline
\end{tabular}
\caption{Description of different datasets regarding LLM4AD.}
\label{fig:different-datasets}
\end{table*}

\subsection{Question Answering}

Question-Answering is an important task that has a wide range of applications in intelligent transportation, assisted driving, and autonomous vehicles~\cite{Xu_2021_CVPR,xu2021sutdtrafficqa}. It mainly reflects through different question and answer paradigms, including traditional QA mechanism~\cite{tang2023domain} and more detailed visual QA methods~\cite{xu2023drivegpt4}.
\cite{tang2023domain} constructs the domain knowledge ontology by “chatting” with ChatGPT. It develops a web-based assistant to enable manual supervision and early intervention at runtime and it guarantees the quality of fully automated distillation results. This question-and-answer system enhances the interactivity of the vehicle, transforms the traditional one-way human-machine interface into an interactive communication experience, and might be able to cultivate the user's sense of participation and control. These sophisticated models~\cite{tang2023domain,xu2023drivegpt4}, equipped with the ability to parse, understand, and generate human-like responses, are pivotal in real-time information processing and provision. They design comprehensive questions related to the scene, including but not limited to vehicle states, navigation assistance, and understanding of traffic situations.
\cite{yang2023humancentric} provides a human-centered perspective and gives several key insights through different prompt designs to enable LLMs to achieve AD system requirements within the cabin. Dolphins~\cite{ma2023dolphins} enhances reasoning capabilities through the innovative Grounded Chain of Thought (GCoT) process and specifically adapts to the driving domain by building driving-specific command data and command adjustments. LingoQA~\cite{marcu2023lingoqa} develops a QA benchmark and datasets, details are in \ref{Section 3.5} and \ref{Section 4}.
EM-VLM4AD~\cite{gopalkrishnan2024multi} is an efficient, lightweight, multi-frame vision language model for Visual Question Answering in autonomous driving, and it only requires much less memory and floating point operations than DriveLM~\cite{sima2023drivelm}. 
~\cite{nouri2024engineeringsafetyrequirementsautonomous} proposes a prototype of a pipeline of prompts and LLMs that receives an item definition and outputs solutions in the form of safety requirements.
Hybrid Reasoning~\cite{azarafza2024hybrid} uses Large Language Models (LLMs) with inputs from image-detected objects and sensor data, including parameters like object distance, car speed, direction, and location, to generate precise brake and speed control values based on weather conditions.
TransGPT~\cite{wang2024transgptmultimodalgenerativepretrained} is a novel large language model for the transportation domain that comes in two variants—TransGPT-SM for single-modal data and TransGPT-MM for multi-modal data—designed to enhance traffic analysis and modeling by generating synthetic traffic scenarios, explaining traffic phenomena, answering traffic-related questions, offering recommendations, and creating comprehensive traffic reports.

\textbf{Metric:}

In terms of QA tasks, NLP's metric is often used. In DriveFPT4~\cite{xu2023drivegpt4} and EM-VLM4AD~\cite{gopalkrishnan2024multi}, it uses BLEU-4~\cite{papineni2002bleu}, METEOR~\cite{banerjee-lavie-2005-meteor}, CIDER\cite{vedantam2015cider} and chatgpt score~\cite{fu2023gptscore}. In \cite{yang2023humancentric}, it adapts accuracy at the individual question level and the command level which includes some sub-questions.

\subsection{Generation}
In the realm of ``generation" task, large language models leverage their advanced knowledge-base and generative capabilities to create realistic driving videos or intricate driving scenarios under specific environmental factors~\cite{khachatryan2023text2videozero,luo2023videofusion}. This approach offers revolutionary solutions to the challenges of data collection and labeling for autonomous driving, also constructing a safe and easily controllable setting for testing and validating the decision boundaries of autonomous driving systems. Moreover, by simulating a variety of driving situations and emergency conditions, the generated content becomes a crucial resource for refining and enriching the emergency response strategies of autonomous driving systems.

The common generative models include the Variational Auto-Encoder(VAE)~\cite{kingma2022autoencoding}, Generative Adversarial Network(GAN)~\cite{goodfellow2014generative}, Normalizing Flow(Flow)\cite{rezende2016variational}, and Denoising Diffusion Probabilistic Model(Diffusion)\cite{ho2020denoising}. With diffusion models have recently achieved great success in text-to-image~\cite{ronneberger2015unet,rombach2021highresolution,ramesh2022hierarchical}, some research has begun to study using diffusion models to generate autonomous driving images or videos. DriveDreamer~\cite{wang2023drivedreamer} is a world model derived from real-world driving scenarios. It uses text, initial image, HDmap, and 3Dbox as input, then 
generates high-quality driving videos and reasonable driving policies. Similarly, Driving Diffusion~\cite{li2023drivingdiffusion} adopts a 3D layout as a control signal to generate realistic multi-view videos. GAIA-1~\cite{hu2023gaia} leverages video, text, and action inputs to generate traffic scenarios, environmental elements, and potential risks. In these methods, text encoder both adopt CLIP~\cite{radford2021learning} which has a better alignment between image and text. In addition to generating autonomous driving videos, traffic scenes can also be generated. CTG++~\cite{zhong2023languageguided} is a scene-level diffusion model that can generate realistic and controllable traffic. It leverages LLMs for translating a user query into a differentiable loss function and use a diffusion model to transform the loss function into realistic, query compliant trajectories. MagicDrive~\cite{gao2023magicdrive}  generates highly realistic images, exploiting geometric information from 3D annotations by independently encoding road maps, object boxes, and camera parameters for precise, geometry-guided synthesis. This approach effectively solves the challenge of multi-camera view consistency. Although it achieves better performance in terms of generation fidelity compared to BEVGen~\cite{swerdlow2023street} and BEVControl~\cite{yang2023bevcontrol}, it also faces huge challenges in some complex scenes, such as night views and unseen weather conditions. ADriver-I~\cite{jia2023adriveri} combines Multimodal Large Language Models(MLLM) and Video Diffusion Model(VDM) to predict the control signal of current frame and the future frames. It shows impressive performance on nuScenes and their private datasets. However, MLLM and VDM are trained separately, which fails to optimize jointly. Driving into the Future ~\cite{wang2023driving} develops a multiview world model, named Drive-WM, which is capable of generating high-quality, controllable, and consistent multi-view videos in autonomous driving scenes. It explores the potential application of the world model in end-to-end planning for autonomous driving.
ChatScene~\cite{zhang2024chatscene} designs an LLM-based agent that generates and simulates challenging safety-critical scenarios in CARLA, improving the collision avoidance capabilities and robustness of autonomous vehicles. REvolve~\cite{hazra2024revolve} is an evolutionary framework utilizing GPT-4 to generate and refine reward functions for autonomous driving through human feedback. 
The reward function is used for RL, and the score is achieved closely by human driving standards. GenAD~\cite{yang2024generalized} is a large-scale video prediction model for autonomous driving that uses extensive web-sourced data and novel temporal reasoning blocks to handle diverse driving scenarios, generalize to unseen datasets in a zero-shot manner, and adapt for action-conditioned prediction or motion planning. DriveDreamer-2 \cite{zhao2024drivedreamer} builds on DriveDreamer with a Large Language Model (LLM), generates customized and high-quality multi-view driving videos by converting user queries into agent trajectories and HDMaps, enhancing training for driving perception methods. ChatSim~\cite{wei2024editable} enable editable photo-realistic 3D driving scene simulations via natural language commands with external digital assets, leverages a large language model agent collaboration framework and novel multi-camera neural radiance field and lighting estimation methods to produce scene-consistent, high-quality outputs. LLM-Assisted Light~\cite{wang2024llmassistedlightleveraginglarge} integrates the human-mimetic reasoning capabilities of LLMs, enabling the signal control algorithm to interpret and respond to complex traffic scenarios with the nuanced judgment typical of human cognition. It developed a closed-loop traffic signal control system, integrating LLMs with a comprehensive suite of interoperable tools. LangProp~\cite{ishida2024langprop} is a framework that iteratively optimizes code generated by large language models (LLMs) using both supervised and reinforcement learning, automatically evaluating code performance, catching exceptions, and feeding results back to the LLM to improve code generation for autonomous driving in CARLA.
These methods explore the customized authentic generations of autonomous driving data. Although these diffusion-based  models achieved good results on video and image-generated metrics, it is still unclear whether they could really be used in closed-loop to really boost the performance of the autonomous driving system.

\textbf{Metric:}

DriveDreamer~\cite{wang2023drivedreamer},  DriveDreamer-2~\cite{zhao2024drivedreamer}, DrivingDiffusion~\cite{li2023drivingdiffusion} and GenAD~\cite{yang2024generalized} use the frame-wise Frechet Inception Distance (FID)~\cite{parmar2022aliased} to evaluate the quality of generated images and the Frechet Video Distance (FVD)~\cite{unterthiner2019accurate} for video quality evaluation. DrivingDiffusion also uses average intersection crossing (mIoU)~\cite{rezatofighi2019generalized} scores for drivable areas and NDS~\cite{yin2021centerbased} for all the object classes by comparing the predicted layout with the ground-truth BEV layout. CTG++~\cite{zhong2023languageguided} following~\cite{xu2022bits,zhong2022guided}, uses the failure rate, Wasserstein distance between normalized histograms of driving profiles, realism deviation (real), and scene-level realism metric (rel real) as metrics.  MagicDrive~\cite{gao2023magicdrive} utilizes segmentation metrics such as Road mIoU and Vehicle mIoU~\cite{Taran_2018}, as well as 3D object detection metrics like mAP\cite{henderson2017endtoend} and NDS~\cite{yin2021centerbased}. ADriver-I~\cite{jia2023adriveri} adapts L1 error of speed and steering angle of the current frame, Frechet Inception Distance(FID), and Frechet Video Distance(FVD) as evaluation indicators. ChatScene~\cite{zhang2024chatscene} provide a thorough evaluation of various scenario generation algorithms. These are assessed based on the collision rate (CR), overall score (OS), and average displacement
error (ADE). REvolve~\cite{hazra2024revolve} adopts fitness scores and episodic steps as metrics in AirSim simulator~\cite{shah2017airsimhighfidelityvisualphysical}.

\subsection{Evaluation \& Benchmark} \label{Section 3.5}
In terms of evaluation, On the Road with GPT-4V~\cite{wen2023road} conducts a comprehensive and multi-faceted evaluation of GPT-4V in various autonomous driving scenarios, including Scenario Understanding, Reasoning, and Acting as a Driver. GPT-4V performs well in scene understanding, intent recognition and driving decision-making. It is good at handling out-of-distribution situations, can accurately assess the intentions of traffic participants, use multi-view images to comprehensively perceive the environment, and accurately identify dynamic interactions between traffic participants. However, GPT-4V still has certain limitations in direction recognition, interpretation of traffic lights, and non-English traffic signs. GPT-4V Takes the Wheel~\cite{huang2023gpt4v} evaluates the potential of GPT-4V for autonomous pedestrian behavior prediction using publicly available datasets. Although GPT-4V has made significant advances in AI capabilities for pedestrian behavior prediction, it still has shortcomings compared with leading traditional domain-specific models.

In terms of benchmark, LMDrive~\cite{shao2023lmdrive} introduces LangAuto(Language-guided Autonomous Driving) CARLA benchmark. It covers various driving scenarios in 8 towns and takes into account 16 different environmental conditions. It contains three tracks: LangAuto track (updates navigation instructions based on location and is divided into sub-tracks of different route lengths), LangAuto-Notice track (adds notification instructions based on navigation instructions), and LangAuto-Sequential track (Combining consecutive instructions into a single long instruction). In addition, LangAuto also uses three main evaluation indicators: route completion, violation score, and driving score to comprehensively evaluate the autonomous driving system's ability to follow instructions and driving safety. LingoQA~\cite{marcu2023lingoqa} developed LingoQA which is used for evaluating video question-answering models for autonomous driving. The evaluation system consists of three main parts: a GPT-4-based evaluation, which determines whether the model's answers are consistent with human answers; and the Lingo-Judge metric, which evaluates the model using a trained text classifier called Lingo-Judge Accuracy of answers; and correlation analysis with human ratings. This analysis involves multiple human annotators rating responses to 17 different models on a scale of 0 to 1, which are interpreted as the likelihood that the response accurately solves the problem. Reason2Drive~\cite{nie2023reason2drive} introduces the protocol to measure the correctness of the reasoning chains to resolve semantic ambiguities. The evaluation process includes four key metrics: Reasoning Alignment, which measures the extent of overlap in logical reasoning; Redundancy, aimed at identifying any repetitive steps; Missing Step, which focuses on pinpointing any crucial steps that are absent but necessary for problem-solving; and Strict Reason, which evaluates scenarios involving visual elements. LaMPilot~\cite{ma2023lampilot} is an  benchmark test used to evaluate the instruction execution capabilities of autonomous vehicles, including three parts: a simulator, a data set, and an evaluator. It employs Python Language Model Programs (LMPs) to interpret human-annotated driving instructions and execute and evaluate them within its framework.~\cite{tanahashi2023evaluation} evaluated the two core capabilities of large language models (LLMs) in the field of autonomous driving: first, the spatial awareness decision-making ability, that is, LLMs can accurately identify the spatial layout based on coordinate information; second, the ability to follow traffic rules to ensure that LLMs Ability to strictly abide by traffic laws while driving.
~\cite{tang2024testing} tests three OpenAI LLMs and several other LLMs on UK Driving Theory Test Practice Questions and Answers, and only GPT-4o passed the test, indicating that the performance of LLMs still needs to be further improved. ~\cite{kong2024superalignment} has developed an LLM-based safe autonomous driving framework, which evaluates and enhances the performance of existing LLM-AD methods in driving safety, sensitive data usage, token consumption and alignment scenarios by integrating security assessment agents. DriveSim~\cite{sreeram2024probing} is a specialized simulator that creates diverse driving scenarios to test and benchmark MLLMs’ understanding and reasoning of real-world driving scenes from a fixed in-car camera perspective. OmniDrive~\cite{wang2024omnidrive} introduces a comprehensive benchmark for visual question-answering tasks related to 3D driving, ensuring strong alignment between agent models and driving tasks through scene description, traffic regulation, 3D grounding, counterfactual reasoning, decision-making, and planning. AIDE~\cite{liang2024aide} proposes an automatic data engine design paradigm, which features automatic data query and labeling using VLM and continuous learning with pseudo-labels. It also introduces a new benchmark to evaluate such automatic data engines for self-driving car perception, providing comprehensive insights across multiple paradigms including open vocabulary detection, semi-supervised learning, and continuous learning. ELM~\cite{zhou2024embodied} is proposed to understand driving scenes over long-scope space and time, showing promising generalization performance in handling complex driving scenarios. LimSim++~\cite{fu2024limsim} introduces an open-source evaluation platform for (M)LLM in autonomous driving, supporting scenario understanding,
decision-making, and evaluation systems.

\section{Datasets in LLM4AD} \label{Section 4}

Traditional datasets such as nuScenes dataset~\cite{nuscenes2019,fong2021panoptic} lack action description~\cite{lu2024activead}, detailed caption, and question-answering pairs which are used to interact with LLMs. 
The BDD-x~\cite{kim2018textual}, Rank2Tell~\cite{sachdeva2023rank2tell}, DriveLM~\cite{sima2023drivelm},  DRAMA~\cite{malla2023drama}, NuPrompt~\cite{wu2023language} and NuScenes-QA~\cite{qian2023nuscenes} datasets represent key developments in LLM4AD research, each bringing unique contributions to understanding agent behaviors and urban traffic dynamics through extensive, diverse, and situation-rich annotations. 
We give a summary of each dataset in Table~\ref{fig:different-datasets}. We give detailed descriptions below.

\textbf{BDD-X Dataset}~\cite{kim2018textual}: With over 77 hours of diverse driving conditions captured in 6,970 videos, this dataset is a collection of real-world driving behaviors, each annotated with descriptions and explanations. It includes 26K activities across 8.4M frames and thus provides a resource for understanding and predicting driver behaviors across different conditions.

\textbf{Honda Research Institute-Advice Dataset (HAD)}~\cite{kim2019CVPR}: HAD offers 30 hours of driving video data paired with natural language advice and videos integrated with can-bus signal data. The advice includes Goal-oriented advice(top-down signal) which is designed to guide the vehicle in a navigation task and Stimulus-driven advice(bottom-up signal) which highlights specific visual cues that the user expects the vehicle controller to actively focus on.

\textbf{Talk2Car}~\cite{Deruyttere_2019}: The Talk2Car dataset contains 11959 commands for the 850 videos of the nuScenes~\cite{nuscenes2019,fong2021panoptic} training set as 3D bounding box annotations. Of the commands, 55.94\% were from videos recorded in Boston, while 44.06\% were from Singapore. On average, each command contains 11.01 words, which includes 2.32 nouns, 2.29 verbs, and 0.62 adjectives. Typically, there are about 14.07 commands in every video. It is a object referral dataset that contains commands written in natural language for self-driving cars.

\textbf{DriveLM Dataset}~\cite{sima2023drivelm}: This dataset integrates human-like reasoning into autonomous driving systems, enhancing Perception, Prediction, and Planning (P3). It employs a ``Graph-of-Thought" structure, encouraging a futuristic approach through ``What if" scenarios, thereby promoting advanced, logic-based reasoning and decision-making mechanisms in driving systems.

\textbf{DRAMA Dataset}~\cite{malla2023drama}: Collected from Tokyo's streets, it includes 17,785 scenario clips captured using the video camera, each clipped to 2 seconds in duration. It contains different annotations: Video-level Q/A, Object-level Q/A, Risk object bounding box, Free-form caption, separate labels for ego-car intention, scene classifier, and suggestions to the driver. 

\textbf{Rank2Tell Dataset}~\cite{sachdeva2023rank2tell}: It is captured from a moving vehicle on highly interactive traffic scenes in the San Francisco Bay Area. It includes 116 clips (~20s each) of 10FPS captured using an instrumented vehicle equipped with three Point Grey Grasshopper video cameras with a resolution of 1920 × 1200 pixels, a Velodyne HDL-64E S2 LiDAR sensor, and high precision GPS. The dataset includes Video-level Q/A, Object-level Q/A, LiDAR and 3D bounding boxes (with tracking), Field of view from 3 cameras (stitched), important object bounding boxes (multiple important objects per frame with multiple levels of importance- High, Medium, Low), free-form captions (multiple captions per object for multiple objects), ego-car intention.

\textbf{NuPrompt Dataset}~\cite{wu2023language}: It represents an expansion of the nuScenes dataset, enriched with annotated language prompts specifically designed for driving scenes. This dataset includes 35,367 language prompts for 3D objects, averaging 5.3 instances per object. This annotation enhances the dataset's practicality in autonomous driving testing and training, particularly in complex scenarios requiring linguistic processing and comprehension. 

\textbf{NuScenes-QA dataset}~\cite{qian2023nuscenes}: It is a dataset in autonomous driving, containing 459,941 question-answer pairs from 34,149 distinct visual scenes. They are partitioned into 376,604 questions from 28,130 scenes for training, and 83,337 from 6,019 scenes for testing. NuScenes-QA showcases a wide array of question lengths, reflecting different complexity levels, making it challenging for AI models. Beyond sheer numbers, the dataset ensures a balanced range of question types and categories, from identifying objects to assessing their behavior, such as whether they are moving or parked. This design inhibits the model's tendency to be biased or rely on linguistic shortcuts.

\textbf{Reason2Drive}~\cite{nie2023reason2drive}: It consists of nuScenes, Waymo, and ONCE datasets with 600,000 video-text pairs labeled by humans and GPT-4. It provides a detailed representation of the driving scene through a unique automatic annotation mode, capturing various elements such as object types, visual and kinematic attributes, and their relationship to the ego vehicle. It has been enhanced with GPT-4 to include complex question-answer pairs and detailed reasoning narratives.

\textbf{LingoQA}~\cite{marcu2023lingoqa}: This dataset is a large-scale, a diverse collection for autonomous driving, containing approximately 419,000 question-answer pairs, covering both action and scenery subsets. It provides rich information about driving behavior, environmental perception, and road conditions through high-quality videos and detailed annotations. It features complex questions and free-form answers, leveraging GPT-3.5/4 to enhance the diversity and depth of content. The driving capabilities covered include actions, reasons, attention, recognition, positioning, etc., which are particularly suitable for improving the understanding and decision-making capabilities of autonomous driving systems.

\textbf{NuInstruct}~\cite{ding2024holistic}: It is a  dataset featuring 91K multi-view video-QA pairs spanning 17 subtasks, each requiring comprehensive information such as temporal, multi-view, and spatial data, thereby significantly raising the complexity of the challenges. It developed a SQL-based method that automatically generates instruction-response pairs, inspired by the logical progression of human decision-making.

\textbf{OpenDV-2K}~\cite{yang2024generalized}: This dataset is a large-scale multimodal dataset for autonomous driving, comprising 2059 hours of curated driving videos, including 1747 hours from YouTube and 312 hours from public datasets, with automatically generated language annotations to support generalized video prediction model training.






\section{Conclusion} 
In this paper, we have provided a comprehensive survey on LLM4AD. We classify and introduce different applications employing LLMs for autonomous driving and summarize the representative approaches in each category. At the same time, we summarize the latest datasets related to LLM4AD. We will continue to monitor developments in the field and highlight future research directions.

\appendix

\section*{Ethical Statement}
When applying LLMs to the field of autonomous driving, we must deeply consider their potential ethical implications. First, the illusion of the model may cause the vehicle to misunderstand the external environment or traffic conditions, thus causing safety hazards. Second, model discrimination and bias may lead to vehicles making unfair or biased decisions in different environments or when facing different groups.  Additionally, false information and errors in reasoning can cause a vehicle to adopt inappropriate or dangerous driving behaviors. Inductive advice may leave the vehicle vulnerable to external interference or malicious behavior. Finally, privacy leakage is also a serious issue, as vehicles may inadvertently reveal sensitive information about the user or the surrounding environment. To sum up, we strongly recommend that before deploying a large language model to an autonomous driving system, an in-depth and detailed ethical review should be conducted to ensure that its decision-making logic is not only technically accurate but also ethically appropriate. At the same time, we call for following the principles of transparency, responsibility, and fairness to ensure the ethics and safety of technology applications. We call on the entire community to work together to ensure reliable and responsible deployment of autonomous driving technology based on large language models.

\section*{Acknowledgments}
 This work was partly supported by NSFC (92370201, 62222607, 61972250) and Shanghai Municipal Science and Technology Major Project (2021SHZDZX0102).
\bibliographystyle{named}
\bibliography{ijcai23}

\end{document}